# Automating the Deep Space Network Data Systems: A Case Study in Adaptive Anomaly Detection through Agentic AI

Evan J. Chou[1,2,3], Lisa S. Locke[3], Harvey M. Soldan[3]

[1]Electrical and Computer Engineering Department, University of California San Diego, 9500 Gilman Drive, La Jolla, CA 91106, USA

[2]Department of Natural Sciences, Pasadena City College, 1570 E Colorado Blvd, Pasadena, CA, 91106

[3]Jet Propulsion Laboratory, California Institute of Technology, 4800 Oak Grove Dr., Pasadena, CA 91109, USA

**Abstract**

The Deep Space Network (DSN) is NASA's largest network of antenna facilities that generate a large volume of multivariate time-series data. These facilities, situated at Canberra - Australia, Madrid - Spain, and Goldstone - California, contain DSN antennas and transmitters that undergo degradation over long periods of time, which may cause costly disruptions to the data flow and threaten the earth-connection of dozens of spacecraft that rely on the Deep Space Network for their lifeline. The purpose of this study was to experiment with different methods and tools that would be able to assist JPL engineers with directly pinpointing anomalies and DSN equipment degradation through collected data, and continue conducting maintenance and operations of the DSN for future space missions around our universe. As such, we have researched various machine learning techniques and architectures that can fully reconstruct data through predictive analysis, and determine anomalous data entries within real-time datasets through statistical computations and thresholds. On top of the fully trained and tested machine learning models, we have also integrated the use of a reinforcement learning subsystem that classifies identified anomalies based on severity level and a Large Language Model that labels an explanation for each anomalous data entry, all of which can be improved and fine-tuned over time through human feedback/input. Specifically, for the DSN transmitters, we have also implemented a full data pipeline system that connects the data extraction, parsing, and processing workflow all together as there was no coherent program or script for performing these tasks before. Using this data pipeline system, we were able to then also connect the models trained from DSN antenna data, completing the data workflow for DSN anomaly detection. This was all wrapped around and further connected by an agentic AI system, where complex reasoning was utilized to determine the classifications and predictions of anomalous data.

**Keywords:** Deep Space Network; Anomaly Detection; Machine Learning; Artificial Intelligence; Transformers; Multivariate Time-Series Data; Reinforcement Learning with Human Feedback; AI Agents; Large Language Models

**Author Roles:** Evan Chou – Author. Lisa Locke – Principal Investigator (mentor). Harvey Soldan – Group Supervisor (co-mentor)

# 1 Introduction
## 1.1 Deep Space Network Background

The Deep Space Network is the largest telecommunications system that provides crucial information and connections to spacecraft beyond Earth. This large system plays a pivotal role in delivering essential telemetry, command, and control capabilities to spacecraft and international space assets situated beyond the confines of our solar system. Two of the main components of the entire Deep Space Network are the antenna and radio frequency (RF) systems, which are essential for transmission and reception of RF signals between DSN communications ground sites on Earth and all our spacecraft in the universe. The RF antennas of the DSN operates on the physical principles of electromagnetism, where RF signals generate an electromagnetic field that radiates energy into space. In the process, each DSS transmitter converts electrical signals into electromagnetic waves during uplink transmission, and vice versa during downlink reception with receivers.

At a high level, the DSN data pipeline consists of many operations and equipment that collect the data and transmit them to software applications and databases. First, a transmitter equipped with a klystron is used in an antenna to amplify and transmit a specific frequency to a spacecraft, and the spacecraft sends back a slightly altered frequency signal. The receivers on the antennae transmit the returned signal through various feed points and a cryogenic receiver that functions as a low-noise amplifier (LNA). The LNA then initiates a cooling process that reduces the temperature of the signal and amplifies it. Finally, the signal is sent to a computer server that serves as a database storage for monitor data. Almost all the data used in this investigation was collected from this database storage and amplification pipeline.

An ongoing project within DSN maintenance and operations is the transition from outdated JPL transmitters to CEC transmitters, which also involves modifying the cooling system, power system, and the transmitter system itself for new design specifications. Because of the difficulty in switching to CEC transmitters, only some DSN antennas are keeping the JPL transmitters temporarily until over the course of the next few years, all the CEC transmitters will replace the JPL transmitters. Therefore, many of the data that is currently being collected periodically in real time comes from both JPL and CEC transmitter data, depending on which transmitters have already been replaced.

## 1.2  Projects Background

The large multivariate time-series datasets collected from the DSN Performance Analysis, NMCLog, and V3WW systems present many correlational features that are useful for examining long-term dependencies and potential abnormalities during certain time periods or timestamps. With the use of machine learning and statistical algorithms, we can conduct operational track data analysis to determine degradation trends within subsystems and antennas that require maintenance from previous missions and equipment aging.

On top of the data collected from DSN antennas, we are also able to collect data from the transmitters in the form of Microsoft Outlook emails. These emails also present a large multivariate time-series dataset with many correlational features, but the format of the data from the transmitters do not come in a structure, tabular dataset format, which is why another separate program focuses on extracting and parsing the data from those emails. By doing so, any operations analyst would be able to cross-reference the transmitter anomalies with the antenna anomalies and gain valuable insight into the anomalous component(s) of the entire DSN system during specific time frames and both DSS antenna number and specific spacecraft features. After this program is developed and shipped, it can then be used to extract and parse data based on the user's datetime preferences.

Lastly, the need for a fully autonomous anomaly detection system for DSN communications ground systems has been emphasized and proposed within Dr. Lisa Locke's previous work. Because of the difficulty in examining long-term degradation trends just from data visualizations, computational methods via machine learning and autonomy must be developed in order to analyze these trends from a wider datetime range that human users would not be able to recognize right away. This autonomous system can be in the form of a singular program that once executed, is able to detect anomalies and generate discrepancy reports within a datetime range.

## 2  Approach

Instead of using traditional machine learning and statistical algorithms, we can turn to modern deep learning architectures and techniques that allow more complex patterns and trends to be analyzed. There are many different types of neural networks and attention-based transformer architectures that allow us to analyze data over long periods of time, and detect unusually sharp fluctuations within time-series data. Some of these architectures and models work well with long-term dependencies, where it is an essential factor that the anomaly detection system should be able to recognize and identify the slow degradation of the antennas and transmitters over months and years.

Furthermore, we can implement an agentic AI system capable of complex reasoning, analysis, and anomaly detection within a static dataset. By doing so, JPL personnel would have an AI assistant that is able to detect anomalies with precision and accuracy, and generate discrepancy reports so that the user can understand the anomaly and corrective action that can be done. To improve the system's adaptability to new data, human feedback can be incorporated into the reinforcement learning component, where users can instruct the AI assistant about whether a detected anomaly really is an anomaly, and possible explanations and corrective actions that would improve discrepancy report generation.

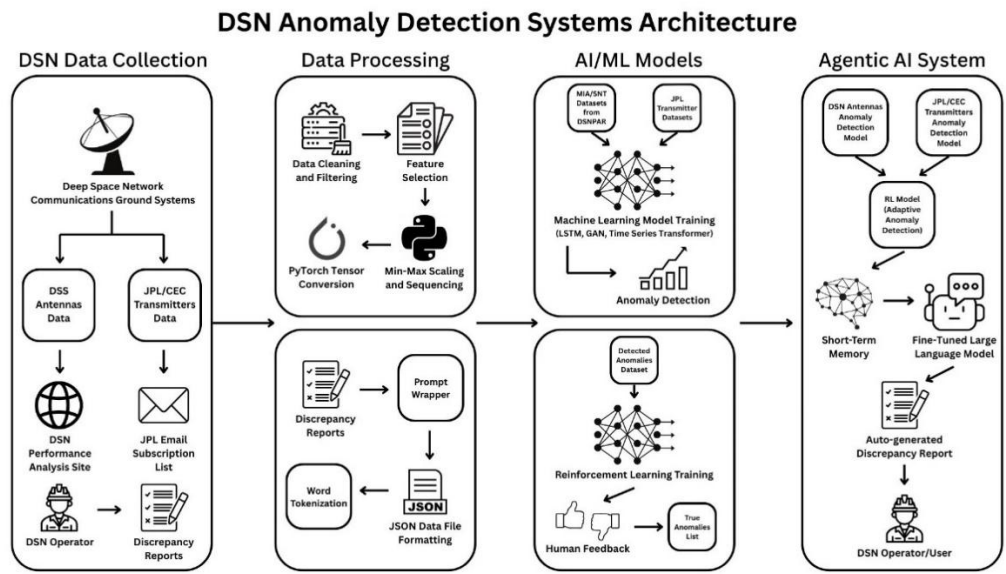

Figure 1: Flow chart of the entire systems architecture for the agentic AI anomaly detection system.

# 3 Data Description and Processing

## 3.1 DSNPAR Datasets

The DSN Performance Analysis site stores DSN monitor data and enables users to query and extract data from an array of subsystems and pipelines. After the monitor data of interest has been queried and extracted, the Excel spreadsheet file is inputted into a MATLAB routine that further formats the dataset. DSNPAR datasets include raw data from MIA, UPLINK, SNT, CONSCAN, and MONOPULSE datasets, and contain hundreds of thousands of data and more than a hundred features, including Power Carrier Number (PCNO), Symbol Signal to Noise Ratio (SSNR), etc.

Previously, a useful program was developed to help make the process of importing datasets from DSNPAR much faster and efficient (Bhargava 2023). With the use of this program, we can query and store data from many months or years for our analysis of the features within DSN antenna and transmitter systems. For most of the machine learning models in this study, we will be using a merged MIA and SNT dataset, where data entries are subsampled every two seconds and can be sorted based on DSS and SCID numbers.

|   | index | t | time | DOY | Date | UPL_FREQ | DL_FREQ | DSS | SCID | AGC | ... | UPL_CMD | UPL_EX | UPL_PWR | WX_HUMID |
|---|---|---|---|---|---|---|---|---|---|---|---|---|---|---|---|
| 0 | 23 | 45658.041667 | 0.041667 | 1 | 45658 | NaN | NaN | 34 | 21 | -93.342 | ... | NaN | NaN | NaN | 26.3 |
| 1 | 26 | 45658.044444 | 0.044444 | 1 | 45658 | NaN | 2.244562e+09 | 34 | 21 | NaN | ... | NaN | NaN | NaN | 25.4 |
| 2 | 27 | 45658.044444 | 0.044444 | 1 | 45658 | NaN | NaN | 34 | 21 | -93.294 | ... | NaN | NaN | NaN | 25.9 |
| 3 | 31 | 45658.047222 | 0.047222 | 1 | 45658 | NaN | NaN | 34 | 21 | -93.305 | ... | NaN | NaN | NaN | 24.9 |
| 4 | 32 | 45658.047222 | 0.047222 | 1 | 45658 | NaN | 2.244562e+09 | 34 | 21 | NaN | ... | NaN | NaN | NaN | 24.7 |
| 5 | 38 | 45658.050000 | 0.050000 | 1 | 45658 | NaN | 2.244562e+09 | 34 | 21 | NaN | ... | NaN | NaN | NaN | 24.8 |
| 6 | 39 | 45658.050000 | 0.050000 | 1 | 45658 | NaN | NaN | 34 | 21 | -93.307 | ... | NaN | NaN | NaN | 24.9 |
| 7 | 48 | 45658.052778 | 0.052778 | 1 | 45658 | NaN | 2.244562e+09 | 34 | 21 | NaN | ... | NaN | NaN | NaN | 23.1 |
| 8 | 49 | 45658.052778 | 0.052778 | 1 | 45658 | NaN | NaN | 34 | 21 | -93.344 | ... | NaN | NaN | NaN | 23.2 |
| 9 | 54 | 45658.055556 | 0.055556 | 1 | 45658 | NaN | NaN | 34 | 21 | -93.331 | ... | NaN | NaN | NaN | 23.9 |

10 rows × 102 columns

Table 1: Preview of the first 10 entries of the combined MIA and SNT dataset from 1/1/25 to 2/10/25.

## 3.2 DSN Transmitter Datasets

Apart from the DSNPAR datasets that come from DSN antenna systems, data from their respective transmitters also provide great insight into anomalies and equipment degradation. By cross-referencing these detected anomalies with those in the DSNPAR datasets, we can further validate and verify the reliability of the anomaly detection system by accounting for more components, sensors, and data sources from the DSN.

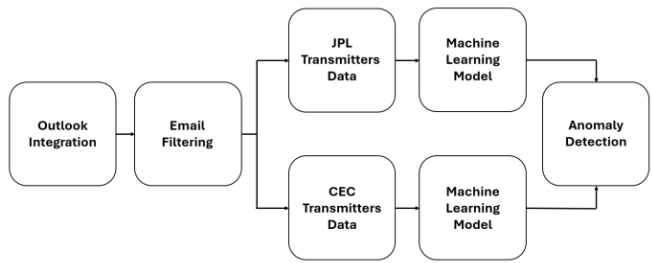

Figure 2: Diagram layout of the entire data pipeline and plan for the DSN transmitter datasets, incorporating Outlook email integration, filtering, and parsing into two separate datasets for different types of DSN transmitters.

DSN transmitter data is collected via an electronic mail system to Microsoft Outlook inboxes, where sets of emails pertaining to different DSS transmitters are emailed periodically to a subscribed list of JPL personnel. To integrate Microsoft Outlook emails within our integrated development environment, we will use the Windows API and Python COM objects library within a Conda environment, which allows us to collect all the user's email inboxes and an array list of email items as long as Microsoft Outlook is opened within the user's operating system. From there, we can then filter by a certain folder, recipient, date, etc. in which separating different datasets by subject line for JPL transmitters and CEC transmitters is necessary for extensive data analysis and operations of all the transmitter data as a whole.

### 3.2.1 JPL Transmitter Datasets

The JPL transmitter datasets are found in emails split into different parts, and within these emails are lines of array-listed numerical data separated by commas. The types of transmitter datasets vary in band-band number pairs, where bands may include S, X, or I bands, and band numbers include either sx20 or t20k. In order to parse all the data contained within these parameters and lines, we must hard-code the executable program that returns the email data into manipulatable Excel or CSV file formats, where each different band and band number pair has their own output file. We are required to separate these pairs into different files due to varying parameters and features, meaning we cannot concatenate them altogether without investigating which datasets and variables correlate with each other. Moreover, further investigation into these emails indicate that some of the emails contain little to no data entries, in which we will simply omit that data from the machine learning training process. After fully formatting data from certain dates of operations, we can then proceed to conduct feature selection and exploratory data analysis to determine which parameters are highly correlated with each other and are likely to have apparent anomalies.

| Variable Names | Non-Null Count | Data Type |
|---|---|---|
| datetime | 1328 | datetime64 |
| dss | 1328 | float64 |
| forward_power | 1328 | float64 |
| reverse_power | 1328 | float64 |
| drive_power | 1328 | float64 |
| exciter_power | 1328 | float64 |
| gain_slope | 1328 | float64 |
| running_time | 1328 | float64 |
| load_t_in_raw | 1328 | float64 |
| load_t_out_raw | 1328 | float64 |
| coll_t_out_raw | 1328 | float64 |
| vac_ion_v | 1328 | float64 |
| vac_ion_on_off | 1328 | float64 |
| fill_air_tach | 1328 | float64 |
| … | … | … |

Table 2: Information from some of the parameters, with their respective non-null values and data types.

Table 3: Preview of the first ten data entries in the JPL transmitter dataset from February 24, 2025. Only eight features are previewed in this screenshot.

### 3.2.2 CEC Transmitter Datasets

Unlike the JPL transmitter datasets, the CEC transmitter datasets come in the form of tar.gz zip files that each contain one large CSV file with a slightly different set of parameters than those in JPL transmitter datasets. Upon further extraction and examination into these CSV files, we noticed that many of these sets of parameters differ from one another, as some have more parameters than others. This led us to investigate the correlation between the CEC transmitters and the specific DSS antenna, where we found that more parameters were seen in the 70-meter antennas compared to the 34-meter antennas. Furthermore, for the purposes of this investigation and machine learning model training, we will simply use features that are similar to forward power (in kW) and body current, while only choosing one classification of transmitter equipment (Agilent, TXC, etc.).

### 3.2.3 Future Work

In the long term, having Microsoft Outlook integration does not extract monitor data and perform real-time anomaly detection. The current program extracts static datasets from emails that are sent daily, but it does not allow for the continuous input of data that is crucial for detecting anomalies in real-time. One of the ways to automate this data collection system and anomaly detection in real-time is by having the transmitter microcontroller directly send the data to the executable program or relational database server, where it will analyze the data for anomalies and send automated discrepancy reports to the same subscribed list of JPL personnel.

## 3.3 Feature Selection

Before training machine learning models based on all the data from the antennas and transmitters, extensive feature selection must be conducted in order to take the most correlated parameters that would be more likely to statistically represent relationships with other parameters and have clear anomalies seen in data visualizations. One of these feature selection methods is Principal Component Analysis (PCA), which is a dimensionality reduction technique that reduces the number of features in a dataset while preserving as much information as possible. PCA isn't a traditional feature selection method; instead of choosing a subset of existing features, it creates new features called principal components which are ranked by how much variance they capture. We experimented with this technique on the MIA dataset, since within that data source, most of the features are highly correlated with one another. Through further analysis alongside data visualizations, we deduced that the variables that represent the highest covariance when it comes to fluctuating anomalous points are SSNR and PCNO data, which we will be experimenting with for machine learning.

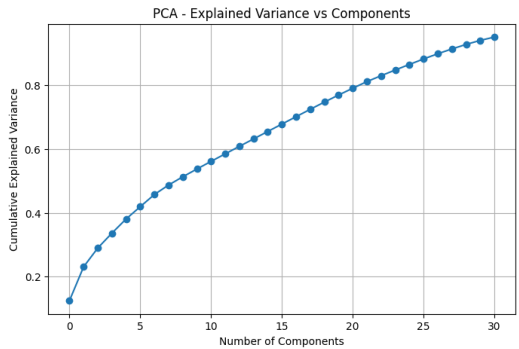

Figure 3: Logarithmic graph of all the principal components and their cumulative explained variance.

Figure 4: Outputted results from PCA, explaining 95% of the variance of different features within the MIA dataset.

## 3.4 AI/ML Data Processing

### 3.4.1 Data Normalization

To normalize our data and prepare it for both outlier filtering and model training, we used the Min-Max Scaler method from the Scikit-Learn preprocessing library. Min-Max Scaler is a normalization technique that transforms features by scaling them into a given range, typically from 0 to 1. We use this method for data normalization because our machine learning models are sensitive to feature scaling, meaning that we must ensure all features contribute equally to distance-based calculations. This is especially the case for our MIA and SNT datasets, where anomalous points can easily stand out in comparison to the tracks that are typically normalized to a common range value. Furthermore, this normalization process will speed up convergence in gradient-based methods for machine learning. However, Min-Max Scaler can be sensitive to outliers that skew the entire scale, and is not ideal for sparse features or when feature distributions are very skewed.

### 3.4.2 Outlier Filtering

Outlier filtering can be a necessary step to take before model training, as it prevents our deep learning weights and biases from overfitting based on the statistically tested outlier/anomalous data. We investigated the use of Isolation Forests, which is an ensemble method and unsupervised machine learning algorithm that isolates observations by randomly selecting a feature and then randomly selecting a split value between the maximum and minimum values of the selected feature (Scikit-Learn). Outliers and anomalies are found through random, recursive partitioning where a sample's path length averaged over a forest of random trees determines whether it is likely to be an outlier. At the final score, an aggregate of depths from all trees classifies data points as anomalies based on a provided contamination parameter that the user sets. Samples that reach shorter tree levels are likely to be anomalies because they are isolated easily.

### 3.4.3 Train/Test Datasets

Many machine learning models require us to convert our timestamps into sequences, as time-series models do not train based on datetime or timestamps within the data. Instead, they take in inputs that are in sequential parts, and the model trains on how those trends are seen from sequence to sequence. Afterwards, we can then split our dataset into train and test sets, and prepare them for tensor conversions and data loading for machine learning model training.

For all our machine learning models and processing, we will use PyTorch, a versatile library for AI/ML that is similar to built-in object-oriented programming (OOP) principles and paradigms. Specifically for deep learning models within the Python IDE, all our data must be converted into tensors of numerical data, instead of using a Pandas dataframe or NumPy vectors. Tensors are essentially a multidimensional vector matrix comprising of three or more dimensions, which makes them similar to arrays and lists data structures by conventional OOP. Many deep learning algorithms in PyTorch require tensors because of accelerated GPU computing, where tensors can be easily transferred for massively parallel operations and distributed training on GPUs. To perform this operation, we used the tensor() function, which converts all our train/test variables to tensors of float32 data type. From there, we put the converted X and y train/test tensors into their respective tensor datasets, and inserted them into a data loader that allows them to be trained on batches of data during every training pass.

## 3.5 Natural Language Processing

Natural Language Processing (NLP) is a subfield of AI and linguistics that focuses on enabling computers and machine learning models to understand, interpret, generate, and interact with natural human language. In simpler terms, NLP is the bridge between human language and machine understanding, where computers learn to understand and generate language, along with its linguistical meanings.

### 3.5.1 Discrepancy Reports

Discrepancy Reports (DR) for the DSN are filed every time an operator finds an anomalous point within real-time monitoring and visualizations, which contain key information regarding the status and descriptions of the discrepancies found. These reports can be used for training an LLM to be able to reasonably determine a course of action and potential cause of a discrepancy, especially when given a user input or prompt wrapper that lists some of the data that caused the anomaly to be found and other situational specifications from the discrepancy. By doing so, the LLM would be able to accurately predict its outputs through probabilistic forecasting, as it is likely that certain discrepancies have been filed in the past that the LLM would be able to remember for future unseen discrepancies and possible resolution plans.

Luckily, a large dataset of discrepancy reports in the form of an h5 file was provided by the Machine Learning and Autonomy Group (398J) at NASA JPL, which lists different parameters for each specific section of a discrepancy report in String data types. With this crucial data, we can then train our chosen pre-trained LLM and adapt it to generate DSN discrepancy reports, listing all of the key information and diagonostics for anomalies and comparing them to previous instances of similar anomalies.

| | SPACECRAFT_ID | GROUND_ANTENNA_ORIG_NUM | GROUND_ANTENNA_CLEAN_NUM | GROUND_ANTENNA_ID | DESCRIPTION_TXT | CORRECTIVE_ACTION_TXT |
|---|---|---|---|---|---|---|
| 0 | 108.0 | 20.0 | 36.0 | 4271.0 | DSS-36: MDS Monitor servers 1 & 4 were intermi... | None |
| 1 | 108.0 | 65.0 | 65.0 | 219.0 | Receiver unexpectedly out of lock at 10:20:03. | Signal reacquired. Carrier locked at 10:23:47.... |
| 2 | 108.0 | 24.0 | 24.0 | 205.0 | DCD failover occurred at 003450Z. Appears to b... | Investigation commenced. |
| 3 | 108.0 | 36.0 | 36.0 | 4271.0 | Receiver dropped lock at 06:52:21. | Reacquisition performed by the LCO. TLM back i... |
| 4 | 108.0 | 34.0 | 34.0 | 213.0 | TLP: FS LOCK STATUS CHANGED triggered from 083... | Station manually acquired.\r\n |
| 5 | 108.0 | 65.0 | 65.0 | 219.0 | Telemetry data was degraded due to the HEMT-S ... | None |
| 6 | 108.0 | 65.0 | 65.0 | 219.0 | Telemetry data was degraded due to the HEMT-S ... | None |
| 7 | 108.0 | 24.0 | 24.0 | 205.0 | Receiver dropped lock unexpectedly | Reacquired downlink |
| 8 | 108.0 | 65.0 | 65.0 | 219.0 | Unable to provide uplink due to S250W1 red sta... | None |
| 9 | 108.0 | 65.0 | 65.0 | 219.0 | Symbol loop dropped lock unexpectedly | Operator reacquired symbol loop |

Table 4: A preview of the first ten rows of the discrepancy reports dataset, featuring some key information in the reports.

### 3.5.2   Prompt Engineering

Before textual data can be used in our LLM training, proper data formatting and processing must be performed in a different way than numerical data for classification or regression models. Prompt Engineering is the practice of strategically designing inputs/prompts to optimize how a LLM generates outputs/responses. One prompt engineering technique is to use prompt wrappers for multivariate datasets, which is a structured way to format inputs dynamically so they are optimized for LLM interactions. This method helps standardize queries for fine-tuning LLMs, which essentially makes LLMs much more predictable and reproducible.

In order to have our LLM take into account all quantitative and qualitative data, we wrote a prompt wrapper that acted as a template for data entries and several features including SPACECRAFT_ID, GROUND_ANTENNA_ORIG_NUM, etc. that were important for specifying the other diagnostics of the anomaly. Each parsed prompt was also followed by a response, which was the CORRECTIVE_ACTION_TXT feature of the discrepancy reports dataset. After all the data prompts and responses have been properly formatted, we sorted them into a JSON file that maps the key-value dictionary pairs into processable training data for LLM fine-tuning. Using the JSON file, we then tokenized all our data using the AutoTokenizer tool from Hugging Face, which prepares the data for LLM fine-tuning.

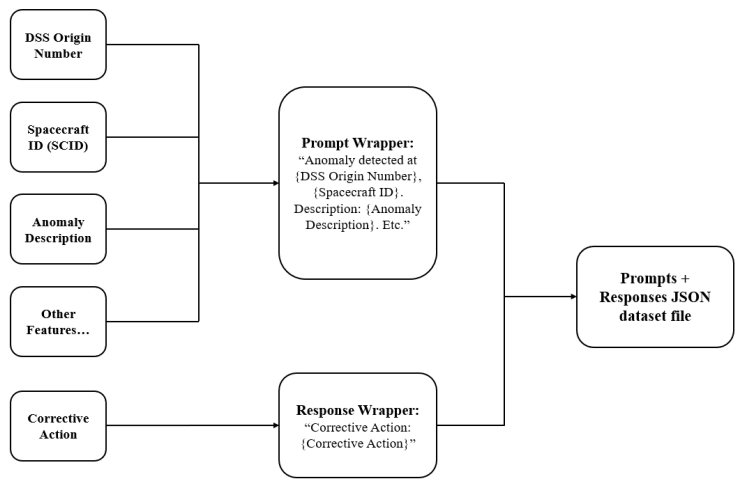

Figure 5: Diagram of the prompt-response structure and data formats into a JSON dataset file.

## 4   Deep Learning-Based Anomaly Detection

Traditionally, classical machine learning algorithms such as linear regression, random forests, etc. have been used to detect anomalies within an unseen dataset. Although they are robust in cases where anomalous data is not present in the training data, classical machine learning algorithms do not capture trends in long-term time dependencies, and especially in sequential modeling where models must understand the statistical significance of the next predicted sequences. As such, we can turn to modern deep learning models and architectures, which involve much more complex mathematics, linear algebra, and matrix operations to change the weights and biases of neural networks. Neural networks differ from classical ML algorithms in the way their architectures are based on nodes, where each node is a weight or bias that is constantly being fine-tuned during training and optimization. Because of its large-scale complexity, this process of changing the weights in a neural network takes lots of time and computing power, therefore requiring the use of accelerated computing via GPU cores[1] and CUDA for massively parallel operations.

One of the most essential parameters of neural networks is the loss function/criterion. For simplicity, we will be using Mean Squared Loss as our criterion, which is a regression loss function that measures the average of the squares of the differences between predicted and actual values. In some cases, we also experimented with Binary Cross-Entropy Loss, a loss function that quantifies the difference between the actual class labels (0 or 1) and the predicted probabilities output by the model. Along with the loss function, backpropagation (backward propagation of errors) is performed during forward and backward passes. This algorithm updates model weights by calculating gradients of the loss function with respect to those parameters. Lastly, the optimizer we will be using is the Adam Optimizer (Adaptive Moment Estimation), which is a gradient-based optimization algorithm that converges fast, is efficient for large datasets, works well with sparse gradients, and has low memory usage.

Throughout the entire phase of deep learning model training, we experimented with different preset parameters before executing the training script. Hyperparameter tuning is the act of manipulating hyperparameters, which are crucial settings that influence how a neural network learns. The most common hyperparameters we used in this phase include the learning rate that determines how much the model adjusts its parameters per iteration, the batch size that defines how many samples the model processes before updating its parameters, and the number of epochs that control how many times the model sees the entire training data. For simplicity purposes, we conducted hyperparameter tuning manually, meaning that we defined each hyperparameter itself and iteratively attempted to improve the accuracy and validation loss of the model.

---

[1] JPL's High Performance Computing allows us to run AI/ML model training on Gattaca 2, which is an onsite physical cluster that has forty-eight compute nodes, four of which have two NVIDIA A100 GPUs and 80GB of VRAM.

## 4.1 Long Short-Term Memory Networks

Long Short-Term Memory (LSTM) networks are a type of Recurrent Neural Network (RNN) capable of learning long-term dependencies, which makes them particularly reliable for time series forecasting and sequential data analysis. LSTMs incorporate a gating mechanism that regulates the flow of information, an issue within RNNs where they aren't able to recognize long-range dependencies due to the vanishing gradient problem. The architecture of an LSTM network contains three gates that control the state and output of the cell state: a forget gate that decides what information should be discarded from the cell state, an input gate that updates the cell state with new information, and an output gate that determines the next hidden gate. In this sense, LSTMs are highly useful for the case of DSN anomaly detection, where we can model and detect long-term degradation trends and update the cell state with new information about how the data is changing over time.

To set up the LSTM and training/testing scripts, we declared an LSTM class that is an instance of a neural network module. In this class, we declared our object constructor with parameters for input size, hidden layer size, number of layers, output size, and a dropout rate of 0.2. LSTM layers are inserted into the neural network architecture with those parameters, followed by a dropout layer that turns off some of the neurons during training to prevent the model from overfitting. The last layer of the LSTM model is the linear layer which acts as the output of the model. We then declared the forward method with an input sequence parameter to indicate the time-series length for each sequence. For this model training, we used a hidden layer size of sixty-four layers, an output size of five variables (for time, DSS, SCID, SSNR, and PCNO), and two LSTM layers. Lastly, we used the Adam Optimizer with a learning rate of 1e-4, and weight decay of 1e-5.

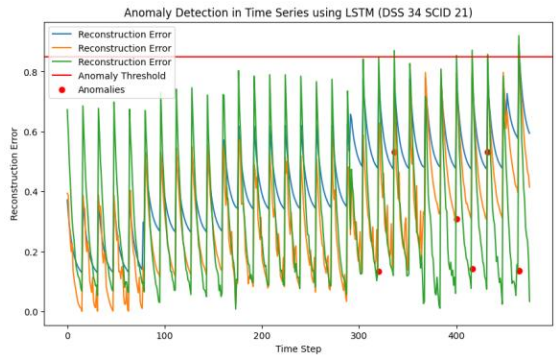

Figure 6: Line plot of LSTM reconstruction errors for DSS 34 and SCID 21.

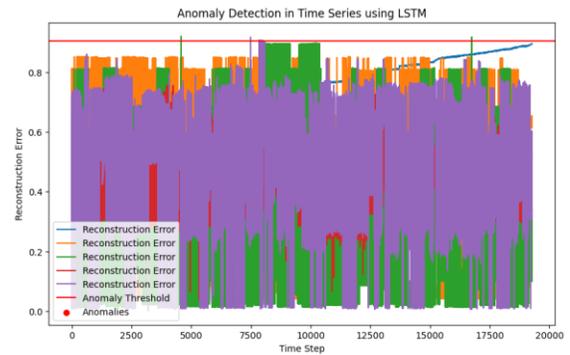

Figure 7: Line plot of LSTM reconstruction errors and anomaly threshold for all DSSs and SCIDs

## 4.2 Generative Adversarial Networks

Another common neural network architecture is Generative Adversarial Networks (GAN), which is a framework primarily used for generative tasks, and works by creating data that mimics a given dataset. GANs have two different neural networks: a Generator that takes random noise as input and learns to generate realistic data samples, and a Discriminator that takes real and generated data as input and learns to distinguish real from fake. These two components work together and are trained simultaneously in a minimax game, where the Generator improves by learning to produce data that increasingly resembles real data, and the Discriminator improves by getting better at telling real from fake. The ideal equilibrium point we hope to converge to is when the Generator produces data so realistic that the Discriminator can't tell the difference.

In our GAN model, on top of using Linear layers, we also experimented with hidden LSTM layers, meaning that we built a hybrid GAN-LSTM model that leverages Generator and Discriminator game training techniques with temporal dependencies and time-series forecasting abilities from LSTM layers. The LSTM layers help generate smooth, realistic temporal transitions in time-series where random noise can be generated, which further improves Discriminator behavior by allowing it to evaluate sequence-level realism, not just point-wise features. Furthermore, because we are running model training specifically on CUDA and GPUs[2], we implemented a series of stability features, including weight initializations, learning rate balancing, and gradient clipping.

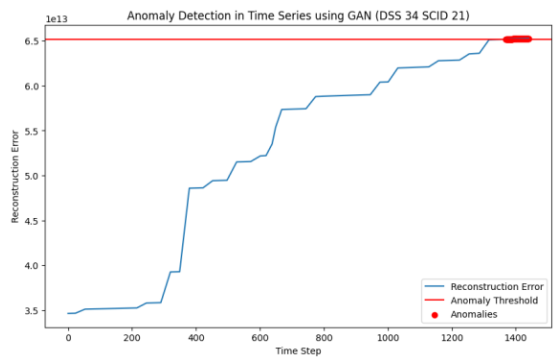

Figure 8: Line plot of GAN reconstruction errors for DSS 34 SCID 21.

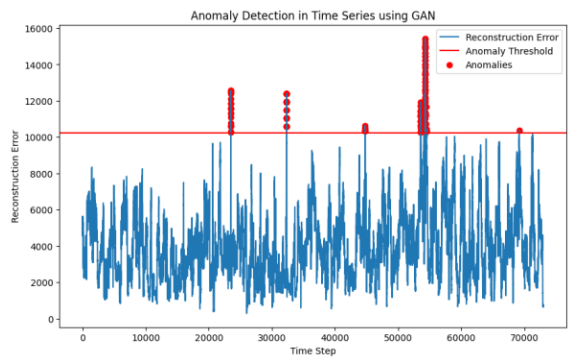

Figure 9: Line plot of GAN reconstruction errors for all DSSs and SCIDs.

---

[2]Because CUDA runs processes differently from running them on a CPU, training and testing a model with a specific code setup may be changed or altered slightly to account for CUDA accelerated computing. This can be due to how gradient accumulation in CUDA might behave differently with their floating-point behavior.

## 4.3 Time-Series Transformers

Transformers are a type of deep learning architecture that uses self-attention mechanisms to model relationships between elements in a sequence without relying on recurrence. Self-attention allows the model to weigh the importance of each data entry(s) in the input relative to every other entry in the same sequence. As such, transformers allow us to analyze which points in the data are the most important and are likely to be anomalies.

A transformer architecture specifically adapted for multivariate time-series data is the Time-Series Transformer (TST). This type of transformer can learn from sequential data with temporal dependencies, and they can also handle long-term dependencies and multivariate sequences as good as LSTMs. In comparison to RNNs and LSTMs, TSTs uses the self-attention mechanism to directly attend to all time sequences which enables better modeling of global patterns and time dependencies. Our TST model was built with input embeddings for the input vector of features, positional encoding for time information, transformer encoder and decoder, and a forecast head that outputs future values or anomaly scores. By taking into account positional encoding in time-series, we can output the uniquely identifiable points that are of upmost importance to analyze.

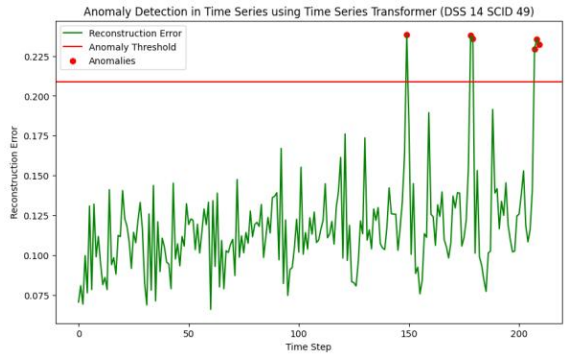

Figure 10: Line plot of TST reconstruction errors for DSS 14 and SCID 49.

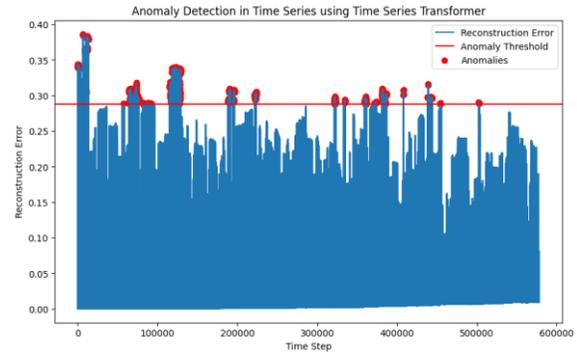

Figure 11: Line plot of TST reconstruction errors for all DSSs and SCIDs.

## 4.4 Results

Based on all the models' detected anomalies within the MIA and SNT dataset, there is much more that needs to be researched and investigated to ensure that the model isn't marking certain data points as false anomalies. Many timestamps and points outputted by the models were verified as false positives, which could potentially be a problem of overgeneralization due to poor hyperparameter tuning. Furthermore, more false positives were found if we slightly lowered the anomaly threshold in the statistical analysis for anomalies, which prompts us to investigate a desirable anomaly threshold that would not be too overgeneralized but still detect anomalies with precision and accuracy. We can also observe that anomalies are more commonly found towards the end of the test dataset sequence, which is unusually expected based on the linear behavior of the models and their reconstruction errors over time steps.

However, through performance analyses between the models and their reconstruction errors, we can infer that GANs perform much more smoothly in comparison to LSTM and TST based on Figure 8, where a smoother and more linear line is plotted. Figure 10 presents lots of noisy reconstruction errors and data, however, anomalous points are much more prevalent and clearer to spot. Comparatively, Figure 6 has structured reconstruction error positioning in certain time step intervals, with higher fluctuated peaks being anomalous points. In this sense, we can analyze certain use cases for different types of deep learning model architectures, where models that tend to reconstruct noisy data may be more suitable for when noisy data is also present in the original dataset, while models that reconstruct smooth data may be suitable for flat linear trends seen during operational track analysis.

## 4.5 Future Work

Within these models in the project, there needs to be more research done regarding hyperparameter tuning, and perhaps using hyperparameter optimization methods such as Grid Search, Random Search, or Bayesian Optimization, which would be able to choose the best sets of hyperparameters that perform the best on the train/test data. This can include changing the number of layers, the activation functions, learning rate, number of epochs, etc. through automated hyperparameter optimization methods. Regularization can also be performed to improve generalization, which can be done by tuning the dropout rate or implementing early stopping during training, ensuring that the model does not learn too well. Furthermore, improvements and extensive re-training of all AI/ML models are necessary to ensure they are adaptable and effective in the long term. This can mean using years of data to train the time-series models to recognize longer-term dependencies over the years, which would require lots of more resources and computing power to allow model training to take place.

# 5 Adaptive Anomaly Classification

After having a list of detected anomalies, we can also perform cross-validation with previous anomalous points in the past and allow a different model to learn from these instances over time. A dataset provided by Dr. Lisa Locke presents different parameters that correspond to the factors that are taken into consideration when deciding whether a detected anomaly is a true anomaly. Some of these parameters include weather, wind, and the respective value for the anomalous feature that flagged the data entry as an anomaly.

## 5.1 Reinforcement Learning

Another field of modern machine learning is reinforcement learning (RL), where models learn the most desirable, best behaviors on an iterative reward-based system. These agents learn to make decisions by interacting with the environment in order to maximize cumulative rewards over time. The learning loop for reinforcement learning involves the agent observing the current state, then choosing an action based on a policy, leading to the environment returning to a new state and reward that allows the agent to update its policy to improve future actions. The ideal convergence for RL models is for them to be able to understand what action to take in each state to maximize total future rewards or returns.

RL can be used as a double verification system in DSN anomaly detection because of how anomalies are rare or ambiguous. Through adaptive learning, RL agents can learn to detect new patterns by continuously interacting with the system. Furthermore, sequential decision-making and reward shaping allows the agent to refine its predictions and convergence because of how anomalies may unfold over time. This is especially useful for our project and investigation because there is a limited amount of data available from previously detected anomalies and their respective data value entries, so improving this system iteratively after it is fully shipped will allow for the system to become more accurate and detect less false positives. For our research, we will place an RL agent on top of our deep learning models to act as a verifier that decides whether to confirm, reject, or request more information about the anomaly in question.

### 5.1.1 Q – Learning

The RL algorithm we experimented with is the Q-Learning algorithm, which is one of the most fundamental RL techniques that learns the optimal action-value function, $Q(s,a)$: this function tells an agent the expected future reward of taking action $a$ in state $s$, and following the optimal policy thereafter. This algorithm doesn't require a model from the environment and can learn from random exploration, hence why it is known as a model-free and off-policy algorithm.

Because the classified anomalies dataset is very simple and interpretable, we felt that there was no need for a PyTorch-based neural network to solve a problem that a Q-table handles perfectly. This allowed us to speed up computations and iterations, interpret the Q-values with ease, and most importantly, easily integrate the model with the agentic AI system and LLM. Within our Q-Learning algorithm, we utilize severity levels of anomalies along with human feedback to determine whether a detected anomaly is a true anomaly. Our reward function declares base rewards for correct or incorrect actions based on the feedback it receives, where a point-based system for low, medium, and high severity levels acts as the states within the algorithm. A severity agent class is also declared, with the action and update functions that utilize the Bellman equation for updating the action-value function.

## 5.2 Human Feedback

The main reason why we chose to experiment with reinforcement learning and Q-Learning algorithms is because of the limited availability of the classified true/false anomalies dataset. Through iterative human feedback, the Q-Learning agent can continue to improve and adapt based on what the user inputs as the feedback on its output/response. The way this works is that after an anomaly is detected from the AI/ML models for statistical reconstruction-based anomalies, it is sent to the Q-Learning agent so that it can analyze the data entry and timestamp for several factors, including wind, rain, etc. that help it confirm whether it is a true anomaly or not. From there, users and operators can analyze the predicted outcome of the anomaly and indicate whether they agree or disagree with the model's classification, enabling the model to iteratively improve from the feedback and the rewards it receives by changing its parameters. Furthermore, after the program has been shipped, the Q-Learning agent can continue to improve and adjust over long periods of operational time, providing more adaptability for our entire anomaly detection system.

## 5.3 Future Work

The entire concept of reinforcement learning and how it applies to adaptive anomaly detection is still an uncertainty that came from very preliminary results of the Q-Learning algorithm in this project. For anomaly detection, reinforcement learning might not be a necessary component of the workflow, as a fine-tuned and adaptive LLM or directly improving the deep learning models would be much more beneficial to the system instead of implementing an extra RL agent in the middle of the workflow. In this sense, investigating other different types of adaptive models that can enable us to implement human feedback for iterative improvement will allow us to still implement and simulate the adaptability of the system, while maintaining a cleaner and more robust workflow.

# 6 Agentic AI

A subfield of AI/ML, agentic AI, is a rapidly growing field that encompasses autonomous agents capable of complex reasoning and action calls. These agents can set goals, take actions, react to feedback, and plan and adapt over time, which is very suitable for full autonomy and complex decision-making for DSN anomaly detection. On a passive system, DSN anomaly detection would simply be using ML models to conduct statistical analysis on a set of static data; however, with agentic AI, more cross-validation can be performed to determine true anomalies and generate responses/reports in human-fluent language, and most importantly, in real-time. In this sense, we could essentially prototype an AI assistant that acts as a full validation method for DSN operators and engineers based on multiple data types and sources.

## 6.1 Large Language Model

Large Language Model (LLM) is a type of transformer capable of generating textual output at human fluency, in which its weights, biases, and parameters can be fine-tuned based on the user's needs. When a user queries an input prompt into the LLM, each token of the prompt is given a uniquely computed vector representation, in which each token can then be passed into the LLM where the model will analyze the probabilistic forecasting of the next token in the sequence and output it.

Using the Discrepancy Reports that were processed into a JSON file, we utilized a Large Language Model to extract contextual information and insights from the DRs, enabling it to act as a human-like entity capable of analyzing anomalies and classifying them based on their metadata and anomalous data. We fine-tuned the chosen Mistral-7B-v0.1 model (with 7 billion parameters), and stored it locally through the Ollama platform, which we will be invoking using an API call to integrate the customized LLM with AI agent-based anomaly detection and complex reasoning. Without the use of data workflows to contextualize the LLM's desired outputs, it would simply be a textual model that operates based on its preexisting knowledge base and reports from the fine-tuning process.

Because fine-tuning pre-trained LLMs can lead to security and data privacy issues during training and inference, it is possible for data leakage to occur especially over cloud-hosted models, where confidential data from the DSN or other NASA projects/missions may be at risk for possible data injections. Instead, we locally hosted the LLM on JPL servers alongside HPC, which can prevent further data privacy issues but still perform at a reasonable level as commercial LLM products. We also ensured that the data we would test during inference phases would not be sent over the cloud and remained only within JPL servers and networks.

### 6.1.1 LoRA Fine-Tuning

Low-Rank Adaptation (LoRA) is an LLM fine-tuning method[3] that freezes the pre-trained model weights and injects trainable rank decomposition matrices into each layer of the Transformer architecture, greatly reducing the number of trainable parameters for downstream tasks (Hu 2021). This method is much more cost-efficient and effective for our project scope than other LLM fine-tuning methods, however, it is possible that extensive LoRA performance will cause the LLM to go through catastrophic forgetting, a phenomenon where the pretrained model loses its knowledge during the process, which is undesirable for our anomaly detection system that is built on previously reported instances of DSN anomalies. For our Mistral-7B-v0.1 model, we performed LoRA fine-tuning with the DR dataset prompts and responses, which enabled us to have a customized LLM for generating DRs and allowing the LLM to output similar corrective actions from what it remembers in its short-term memory.

## 6.2 AI Agent Workflow

The most important component of the project is the entire AI agent workflow, which is orchestrated by the LangGraph framework. In each step of the agent workflow, action calls are implemented in order of their reasoning and decision-making. We used a StateGraph in our agentic AI workflow to dictate the anomalies list, logs, messages, decision, feedback, and any other passed data or attributes it receives and manipulates throughout the agent orchestration. Each action is a function call, where we have declared functions for computing anomaly scores, explaining decisions, planning decisions, and incorporating human feedback when a static dataset is imported into the workflow. During the inference step, we invoke an API call to the fine-tuned LLM stored in an Ollama local server, using it as the reasoning agent and automated discrepancy report generator. To integrate multimodal anomaly detection through DSN antennas and transmitter components/datasets, we organized the workflow by bringing in data pipelines designated for each specific data source and then connecting them all to the LLM agent through function calls. By bringing in all the deep learning models, RL model, and LLM by seamlessly connecting them, we are able to orchestrate a fully organized maintenance and operations cycle for DSN communications ground systems anomaly detection, comprising of all the AI/ML models we previously trained and fine-tuned.

## 6.3 Future Work

To continue this investigation, future testing and research must be done to determine the best orchestrated workflow at which information will flow from one to the other. Among many considerations, different types of systems architecture for the entire LangGraph workflow can be manipulated freely depending on the user's preferences, which enables for varying results outputted from the LLM agent. Furthermore, as more changes are made to the DSN and the application's other dependencies, this AI agent workflow can be easily reformatted with additional steps between the workflow or by equipping the agent with more functionality tools. This also includes switching to improved and optimized machine learning models that have benchmarked better performances than the ones trained and used in this investigation, or re-coding logic within each function of the agent so that it better analyzes anomalous data and integrates all the information together before decision-making. Furthermore, because our current AI agent system and workflow is far from being a fully agentic system, real-time data pipelines and autonomy can be implemented straight from the data sources via Apache Kafka or Spark, which allows for the system to analyze real-time monitor data instead of static datasets, operating independently without user intervention unless requested for.

---

[3]LoRA can only be performed and executed when run on CUDA and GPU computing nodes via JPL HPC, as the Hugging Face documentation lists CUDA as a hardware accelerator requirement before fully fine-tuning the LLM. Running LoRA on the CPU will return a Runtime Error.

# 7 Related Work

## 7.1 A Review of Anomaly Detection in Spacecraft Telemetry

Fejjari, Delavault, Camilleri, and Valentino (2025) investigated various machine learning techniques applied to anomaly detection for spacecraft telemetry, including Support Vector Machines, clustering algorithms, and deep learning models such as Convolutional Neural Networks and RNNs. In their work, they emphasized the complexity of telemetry data, which includes high-dimensional, time-series information with intricate interdependencies among parameters, noting that while deep learning models can capture these complex patterns, they often require substantial computational resources and large datasets. However, the most significant insight presented in this research is the importance of incorporating domain knowledge into anomaly detection systems, which includes operational context and expected behavior of spacecraft systems. By understanding these aspects, anomaly detection systems can be more accurately tuned to identify true anomalies while reducing false positives.

## 7.2 Detection and quantification of anomalies in communication networks based on LSTM-ARIMA combined model

Xue, Chen, and Zheng (2022) presented a hybrid approach to anomaly detection in communication networks/systems by integrating LSTM with Autoregressive Integrated Moving Average (ARIMA) models. They proposed a prediction-then-detection framework that operates using an LSTM model to capture non-linear temporal dependencies in data, and then an ARIMA model that handles linear trends and seasonality within the data, ensuring that the training data is cleaned through outlier elimination. The predictions from both models are combined using a residual weighting technique, which balances the strengths of each model to forecast future states of key performance indicators (KPI) without outliers. This investigation resulted in successful performance in predicting KPI values compared to individual models, as the combined approach effectively captured both linear and non-linear patterns in the data which led to more accurate forecasts. The main key insight is that integrating different modeling techniques and algorithms can leverage their individual strengths, resulting in a more robust and accurate anomaly detection system.

# 8 Discussion and Implications

This investigation was formulated by Group 333F, DSN Systems Engineering, and we were tasked with developing and prototyping an autonomous anomaly detection system to predict the performance of antennas and transmitters, identify degradation trends over long periods of operational track analysis, and generate fault diagnostics and discrepancy reports. The end goal of the entire investigation is to have an assistant tool that is able to aid DSN operators and engineers in diagnosing anomalies and ensuring that the DSN antennas and transmitters are operating normally. In doing so, the DSN teams will be able to forecast future degradation trends and equipment replacement needs before encountering a critical situation or emergency where the DSN is failing to receive signals from a spacecraft in our universe.

This spring term, we researched and developed a machine learning-based system for statistical anomaly detection, focusing on the use of agentic AI and deep learning model architectures for complex reasoning and analysis. Our contributions spanned many different phases of the AI/ML and software development cycle, from the data retrieval and processing stage to the fully implemented AI system.

1. Data Retrieval and Processing
    a) Developed a program that extracts and parses DSN transmitter data into manipulatable data frames.
    b) Conducted feature selection, scaling, and outlier filtering for MIA, SNT datasets and JPL, CEC transmitter datasets.
    c) Tokenized DSN discrepancy reports and textual data for cross-referencing past anomalies through an LLM.
2. ML Model Training/Testing
    a) Investigated three different deep learning architectures (LSTM, GAN, TST) for reconstruction-type anomaly detection.
    b) Implemented human feedback into an RL algorithm (Q-Learning) for identifying true positives in detected anomalies.
3. Agentic AI System
    a) Orchestrated an AI agent workflow for complex reasoning and decision-making in determining anomalies.
    b) Utilized a fine-tuned LLM agent for generating discrepancy reports and incorporating human feedback.

# 9 Future Work

## 9.1 Apache Spark Integration

As previously proposed in Section 6.3, real-time data pipelines and automation for a fully agentic system can be accomplished through streaming frameworks such as Apache Spark and Kafka. This framework works well when dealing with large-scale, high-throughput, and real-time data input, enabling the system to utilize data ingestion and feature engineering to automate the entire data collection and analysis steps of the workflow. One of the helpful features within this framework is that it can send alerts to different distributed systems and dashboards, which may work well for human operators to utilize when keeping track of the input data and detected anomalies in real-time. If integrated smoothly into the agentic AI workflow, the anomaly detection process would be completely automated, and human operators may only be required to analyze the AI assistant's auto-generated reports, detailed analysis, and reasoning process for statistically detected anomalies.

Although not commonly used for this purpose, Apache Spark and Kafka also offer machine learning-based anomaly detection tools that can be implemented into the streaming process, which can be tested and validated to determine whether built-in tools within certain software technologies will suffice for the entire anomaly detection process. When determining the usability of these tools, the system must be tested on long periods of data, allowing us to analyze whether the machine learning algorithms provided by the framework can also recognize the long-term degradation trends, and not just a fixed linear trend that is only representative of a short period of operational track analysis.

## 9.2 Transfer Learning

Transfer Learning is an ML technique in which knowledge gained through one task or dataset is used to improve model performance on another related task and/or different dataset. The way this technique works is by using what has been learned in one setting to improve generalization in another setting, which means that transfer learning takes a pre-trained model and apply its knowledge gained in an initial task or data toward a new target task or data. However, transfer learning only works best when all the learning tasks are similar, source and target datasets data distributions do not vary too greatly, and a comparable model can be applied to all tasks (Murel 2024).

In the context of DSN anomaly detection, transfer learning can be employed to adapt our models trained on data from one DSS-SCID pair to detect anomalies in another pair. We can perform the transfer learning approach with our datasets because there are similarities in operational patterns or telemetry structures across different systems, as long as the data (specifically, SSNR and PCNO data) has all been scaled and preprocessed accordingly. By implementing transfer learning into our deep learning models, we are able to not only reduce computational resources and the amount of data required, but we can develop the system to be more efficient and adaptable for each different DSS-SCID pair.

## 10 Conclusion

Over the past seventeen weeks, we researched and developed agentic AI-based anomaly detection systems for DSN communications ground systems, capable of detecting and pinpointing true anomalies. Although all our statistical results were preliminary and the models were not robust to potential false positives and inaccuracies, they gave us insights into how we can conduct further implementation through iterative training and testing for future investigations. We were able to lay a framework and systems architecture for how JPL engineers can analyze DSN antenna and transmitter data through multiple methods and pipelines, all of which are incorporating both AI and modern machine learning techniques to encompass a wider array of cross-referencing information and static dataset sources. Different data sources and models can be integrated into the workflow, allowing JPL engineers to flexibly manipulate and contribute to the system. Overall, this project helped us analyze discrepancies and anomalous timestamps in the DSN, and how we can improve and fine-tune the data systems and pipelines so that a more accurate and precise AI assistant is able to adapt and recognize the slow degradation trends of the Deep Space Network.

## Acknowledgements

I would like to thank Dr. Lisa Locke, Michael Settember, Harvey Soldan, Jason Liao, and NASA JPL Group 333F for guiding me along my research projects and learning journey. Other JPL employees I would like to thank include Dr. Umaa Rebbapragada from 398J and Rob Royce from 192F for providing AI, machine learning, and data science advice throughout the duration of the project. The High Performance Computing resources used in this investigation were provided by funding from the JPL Information and Technology Solutions Directorate. Lastly, I would like to thank the NASA JPL STEM Engagement Office for their continuous support and opportunity to work on the project.

*This research was carried out at the Jet Propulsion Laboratory, California Institute of Technology, and was sponsored by the Student Independent Research Internship Program (SIRI) and the National Aeronautics and Space Administration (80NM0018D0004).*

## References

ADL. (2018, September 3). *An introduction to Q-Learning: reinforcement learning*. freeCodeCamp. https://www.freecodecamp.org/news/an-introduction-to-q-learning-reinforcement-learning-14ac0b4493cc/.

Bhargava, E., Sikorski, A., Soldan, H, & Locke, L. (2023, July 28). *Improving the Deep Space Network Data Pipeline: A Case Study in Data Engineering and Machine Learning-based Anomaly Detection*. NASA Jet Propulsion Laboratory.

Fejjari, A., Delavault, A., Camilleri, R., & Valentino, G. (2025, May 19). *A Review of Anomaly Detection in Spacecraft Telemetry Data*. University of Malta. https://www.mdpi.com/2076-3417/15/10/5653.

Goodfellow, J., Pouget-Abadie, J., et al. (2014, June 10). *Generative Adversarial Nets*. Université de Montréal. https://arxiv.org/pdf/1406.2661.

*High Performance Computing*. (2025, April 11). JPL Information Technology. https://jplit.jpl.nasa.gov/it-services/high-performance-computing.

Hu, E., Shen, Y., et al. (2021, June 17). *LoRA: Low-Rank Adaptation of Large Language Models*. Hugging Face. https://huggingface.co/papers/2106.09685.

Liao, Jason. (n.d.). *DSN Performance Analysis*. DSN Systems Engineering Group 333F. https://webvpn.jpl.nasa.gov/https/dsnpar.jpl.nasa.gov/.

Jiang, A., Sablayrolles, A., et al. (2023, October 10). *Mistral 7B*. Mistral AI. https://arxiv.org/pdf/2310.06825.

Kingma, Diederik & Ba, Jimmy Li. (2017, January 30). *Adam: A Method for Stochastic Optimization*. International Conference for Learning Representations. https://arxiv.org/pdf/1412.6980.

Murel, Jacob & Kavlakoglu, Eda. (2024, February 12). *What is transfer learning?* IBM. https://www.ibm.com/think/topics/transfer-learning.

Sooriyarachchi, Avinash. (2023, August 30). *Efficient Fine-Tuning with LoRA: A Guide to Optimal Parameter Selection for Large Language Models*. Databricks. https://www.databricks.com/blog/efficient-fine-tuning-lora-guide-llms.

Staudemeyer, R., & Morris, E. (2019, September 23). *Understanding LSTM – a tutorial into Long Short-Term Memory Recurrent Neural Networks*. https://arxiv.org/pdf/1909.09586.

*Time Series Transformer*. (n.d.). Hugging Face. https://huggingface.co/docs/transformers/model_doc/time_series_transformer.

Xue, S., Chen, H., & Zheng, X. (2022, June 17). *Detection and quantification of anomalies in communication networks based on LSTM-ARIMA combined model*. Anhui University of Science and Technology. https://pmc.ncbi.nlm.nih.gov/articles/PMC9205417/.